\documentclass[lettersize,journal]{IEEEtran}
\usepackage{amsmath,amsfonts}
\usepackage{algorithmic}
\usepackage{algorithm}
\usepackage{array}
\usepackage[caption=false,font=normalsize,labelfont=sf,textfont=sf]{subfig}
\usepackage{textcomp}
\usepackage{stfloats}
\usepackage{url}
\usepackage{verbatim}
\usepackage{graphicx}
\usepackage{cite}
\usepackage{multicol} 
\usepackage{etoolbox}

\makeatletter
\newcommand\blfootnote[1]{%
  \begingroup
  \renewcommand\thefootnote{}\footnote{#1}%
  \addtocounter{footnote}{-1}%
  \endgroup
}
\makeatother

\begin{document}

\title{From Air to Wear: Personalized 3D Digital Fashion with AR/VR Immersive 3D Sketching}

\author{Ying Zang, Yuanqi Hu, Xinyu Chen, Yuxia Xu, Suhui Wang, Chunan Yu, Lanyun Zhu, Deyi Ji,  Xin Xu, Tianrun Chen*}

\markboth{Journal of \LaTeX\ Class Files,~Vol.~14, No.~8, August~2021}%
{Shell \MakeLowercase{\textit{et al.}}: A Sample Article Using IEEEtran.cls for IEEE Journals}




\twocolumn[{
\renewcommand\twocolumn[1][]{#1}
\maketitle
\begin{center}
\vspace{-1cm}
    \includegraphics[width=1.0\textwidth]{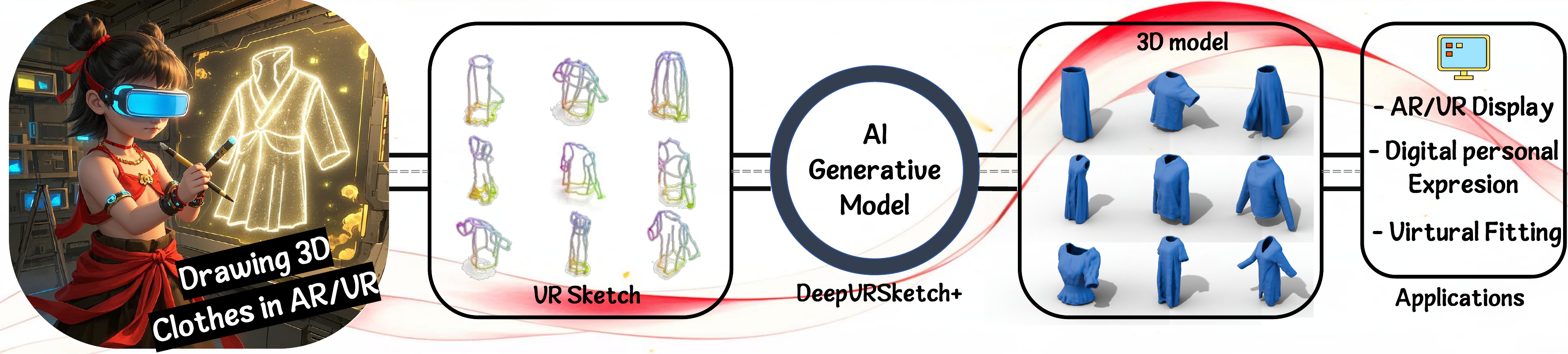}
\end{center}
\noindent

\refstepcounter{figure} 
\footnotesize \textsf{Fig. \thefigure. \hspace{0.3em} 
In this work, we propose a novel method that allows everyday users to create personalized 3D garments by simply sketching in immersive AR/VR environments (in 3D). A carefully designed generative AI model transforms these freehand 3D sketches into realistic, detailed garment models, which can be used for personalized expression in the metaverse, AR/VR visualization, and virtual try-on applications.}
}]

\begin{abstract}
In the era of immersive consumer electronics, such as AR/VR headsets and smart devices, people increasingly seek ways to express their identity through virtual fashion. However, existing 3D garment design tools remain inaccessible to everyday users due to steep technical barriers and limited data. In this work, we introduce a 3D sketch-driven 3D garment generation framework that empowers ordinary users — even those without design experience — to create high-quality digital clothing through simple 3D sketches in AR/VR environments. By combining a conditional diffusion model, a sketch encoder trained in a shared latent space, and an adaptive curriculum learning strategy, our system interprets imprecise, free-hand input and produces realistic, personalized garments. To address the scarcity of training data, we also introduce KO3DClothes, a new dataset of paired 3D garments and user-created sketches. Extensive experiments and user studies confirm that our method significantly outperforms existing baselines in both fidelity and usability, demonstrating its promise for democratized fashion design on next-generation consumer platforms.
\end{abstract}

\begin{IEEEkeywords}
AR/VR, 3D Sketch, Shape-from-X, Content Creation, Metaverse.
\vspace{-0.5cm}
\end{IEEEkeywords}

\blfootnote{This work is supported by the Public Welfare Research Program of Huzhou Science and Technology Bureau (2022GZ01) and ZJU Kunpeng \& Ascend Center of Excellence. (\textit{Corresponding author: Tianrun Chen})}
\blfootnote{Ying Zang, Yuanqi Hu, Xinyu Chen, Yuxia Xu, Suhui Wang and Chunan Yu are with the School of Information Engineering, Huzhou University. (e-mail: 02750@zjhu.edu.cn)}
\blfootnote{Lanyun Zhu is with the Information Systems Technology and Design Pillar, Singapore University of Technology and Design.}
\blfootnote{Deyi Ji is with the School of Information Science and Technology, University of Science and Technology of China}
\blfootnote{Xin Xu is with KOKONI, Moxin (Huzhou) Technology Co., LTD.}
\blfootnote{Tianrun Chen is with the College of Computer Science and Technology, Zhejiang University and KOKONI, Moxin (Huzhou) Technology Co., LTD. (email: tianrun.chen@zju.edu.cn)}
\blfootnote{*Corresponding Author}

\section{Introduction}
\label{sec:intro}
\IEEEPARstart{C}{lothing} is one of the most personal and powerful forms of self-expression \cite{zhang2024mannequin2real, hu2024towards, plazyk2015democratization}. People don’t just wear clothes — they choose them to reflect their identity, taste, and mood. In the digital age, clothing is no longer limited to the physical world. As AR/VR technologies become part of everyday consumer electronics — from headsets and smart mirrors to mobile AR apps — people are spending more and more time in immersive virtual spaces. In these spaces, clothing still matters. Just like in real life, people want to express their identity through what they wear in the virtual world. But unlike the real world, where options are limited by manufacturing and logistics, digital garments can be limitless — as long as we give people the tools to create them.

However, for a long time, fashion design was a privilege reserved for the elite — professional designers with years of training and expensive tools \cite{hopkins2021fashion, arora2017experimental}. Designing even a single 3D garment required mastering complex modeling software and navigating tedious workflows. Ordinary users still don’t have accessible tools to design clothing in 3D. That’s the gap we aim to fill. Just as AR headsets and VR devices are becoming more affordable and user-friendly — turning once high-end tech into everyday consumer products — we believe 3D fashion creation should follow the same path. Therefore, in this work, we develop a fashion design tool that is just as democratized - intuitive, creative, and open to all, with the growing accessibility of AR/VR platforms.

    To make 3D clothing creation accessible to everyone, the first challenge we need to solve is how to simplify the design process. Traditional 3D modeling tools are complex and time-consuming, often requiring years of training and professional software — far beyond the reach of everyday users \cite{gao2022sketchsampler}. But with the rise of consumer-grade AR/VR devices, the way people interact with 3D space is rapidly changing. Here, we innovatively propose to use AR/VR devices as the creation tool.

    AR/VR headsets not only let users view 3D objects in immersive environments, but also allow them to create directly in 3D space. Past researches have shown that with motion controllers or hand tracking, anyone can sketch in the air — drawing curves and shapes as if they were sculpting with invisible tools \cite{gao2022sketchsampler, deering1995holosketch, keefe2001cavepainting, kwan2019mobi3dsketch, xu2018model}. We build on this capability by pairing these 3D sketches with our carefully designed generative AI model, which instantly transforms rough strokes into high-quality 3D garments that match the user’s vision (see main figure). With AI generative model, we removes the need for accurate drawing or manual modeling. Instead, users can focus on expressing ideas, while the AI fills in the structure and detail. We believe sketching is one of the most natural and powerful ways for people to communicate creative ideas. It's quick, intuitive, and nearly everyone can do it \cite{arora2017experimental}.

    The real challenge, though, is making sure the system understands the sketch — especially when it’s loose or imprecise. For beginners, drawing precise lines is often difficult, especially when the clothing has a freely shaped surface, making it challenging to accurately depict. But clothing is complex, often with soft, flowing shapes that are hard to capture in rough lines. While past methods rely on supervised learning or direct regression to match sketches to models \cite{zheng2021pamir, he2023sketch2cloth}, they often struggle when the sketch is unclear. Our goal is to go further: to create a system that’s not only easy to use, but also forgiving — one that helps all users, regardless of skill, generate high-quality, faithful 3D designs.

    To address the challenge, we adopt a generative approach instead of determinstic regression to obtain 3D design. Rather than directly regressing a 3D model from a user’s sketch — a process that often requires precise input — our network treats the sketch as a loose condition. This allows the model to "imagine" a complete and realistic 3D garment that aligns with the user’s intent, even when the input is rough or incomplete. The generative model learns to interpret abstract lines and transform them into high-fidelity 3D shapes with natural structure and flow, lowering the technical barrier for users.

    However, generative models typically rely on large datasets to perform well — and in the 3D fashion domain, data scarcity remains a pressing issue. The most widely used dataset, provided by Zhu et al. \cite{zheng2021pamir}, contains only 1,212 3D garments and lacks paired human-drawn sketches. This makes it difficult to train reliable, generalizable models. As a result, finding ways to fully leverage this limited data becomes a key challenge in making sketch-based generative 3D clothing design truly practical and accessible.

    To tackle the limitations posed by scarce training data, we introduce a three-stage strategy that leverages a shared 3D point cloud-based latent space, curriculum learning, and a newly collected dataset. First, we pre-train a conditional diffusion model on a large-scale, diversified dataset of clothes 3D shapes and point clouds. This diffusion model serves as a strong shape generator, producing high-quality garments from abstract latent features. In the second stage, we freeze the pre-trained diffusion model and train a sketch encoder that maps a hand-drawn 3D sketch into the same latent space. These encoded features are injected into the intermediate layers of the diffusion model to condition the generation process, allowing it to adaptively synthesize garments that reflect the user’s input sketch while preserving shape realism and plausibility. Finally, we jointly fine-tune the entire system — the sketch encoder and the diffusion generator — to further enhance the alignment between user sketches and generated garments. During training, we observed that the model struggled to generalize across diverse sketching styles and garment geometries, especially under limited supervision. To mitigate this, we adopt an adaptive curriculum learning strategy that gradually increases the complexity of training samples, helping the model learn from simpler shapes before progressing to more intricate examples. 
    
    To further alleviate data scarcity, we contribute a new dataset — KO3DClothes — which includes paired 3D garments and human-drawn 3D VR sketches. We select 3D garment models from the DeepFashion3D dataset \cite{zheng2021pamir} and invite 10 non-professional participants to create corresponding sketches using custom VR sketching tools. The resulting dataset contains 969 high-quality paired samples, enriching the available resources for research on sketch-based 3D garment generation and facilitating future work in this direction.

     Through comprehensive experimental validation, our method outperforms existing benchmarks in both model quality and realism, even generating satisfactory results for previously unseen data drawn by novice users. Additionally, we demonstrate that, even with intricate and complex structures, the system can successfully generate accurate 3D models based on detailed 3D drawings from users. In user studies, participants expressed high satisfaction with the generated 3D models, further validating the practicality and effectiveness of our approach. We believe that our work takes a step toward democratizing digital fashion by making 3D garment design intuitive, expressive, and widely accessible for next-generation consumer platforms.

\section{Related Works}
\label{sec:Related works}
\subsection{3D Garment Design with Deep Learning Algorithm.}
    Given the significance of 3D garment design, many methods have been proposed \cite{zhao2024weakly, jiang2020bcnet, zhu2022registeringexplicitimplicithighfidelity, srivastava2022xcloth, bhatnagar2019multi, zhu2020deepfashion3ddatasetbenchmark} to reconstruct or digitize 3D garments from images. Some approaches \cite{saito2019pifu, saito2020pifuhd, zheng2021pamir, xiu2023econ} employ neural implicit representations, such as occupancy fields and Signed Distance Functions (SDF), to construct 3D models of clothed humans from single-view or sparse multi-view images using supervised learning. However, these methods face limitations when it comes to independently modeling garments separate from the body. ReEF \cite{zhu2022registeringexplicitimplicithighfidelity} addresses this issue by employing a technique that independently models garments through the learning of explicit boundary curves and segmentation fields. In contrast, xCloth \cite{srivastava2022xcloth} offers a more efficient representation with the added advantage of generating texture maps. Nonetheless, these approaches still rely on high-quality real-world datasets of clothed humans, which often lack diversity in style and appearance, mainly due to the high cost of acquiring large-scale datasets. Furthermore, the generated garments are usually tightly coupled with the underlying body pose, often resulting in suboptimal surface quality.
    
    Another research direction, inspired by the actual garment creation process, has proposed both analytical \cite{pietroni2022computational} and neural network-based \cite{korosteleva2021generating} methods for procedurally generating unposed 3D garments ready for production. However, these methods rely on complex sewing patterns, which are not intuitive for designers. More recent approaches \cite{corona2021smplicit, su2022deepcloth} have bypassed panel-based generation by using parametric human body templates to generate garment models. However, these methods typically focus on modeling tight-fitting garments and need substantial expertise to create the 3D garment.
    
    In contrast to previous approaches, our method exclusively uses \textbf{VR sketches} as the input modality, allowing novice users to create the 3D garment without worrying about the issue of spatial perception in 3D in 2D and drawing skills. There are some existing sketch-based 3D garment design methods \cite{yu2023surf, chowdhury2022garment, he2023sketch2cloth, bandyopadhyay2023doodle}, but they are limited by the 2D input. The user needs to ``imagine" 3D before drawing, which is hard. Despite the input differences, we also adopt a more complex but effective network configuration and training scheme compared to these previous works to tackle the data scarcity and low output quality issue. We design a multi-stage generative network that, even with limited training data, can generate high-quality 3D shapes while accurately capturing the user’s design intentions.

\subsection{3D Model Generation with Generative Models}
    In recent years, significant progress has been made in 3D shape generation. Many studies have explored various generative models, including Generative Adversarial Networks (GANs), Variational Autoencoders (VAEs), autoregressive models, normalizing flows, and more recently, diffusion models. In this study, we adopt a diffusion model-based network architecture, which has achieved state-of-the-art results in the field of 3D shape generation \cite{cheng2023sdfusion, kong2022diffusion, nam20223d}, capable of producing high-quality and highly detailed 3D shapes.
    
    Additionally, 3D model generation methods based on images and text have also made significant strides in recent years \cite{sanghi2022clip, li2023generative, fu2022shapecrafter, tian2023shapescaffolder,liu2023one,shi2023mvdream}. However, in contrast to traditional approaches, we introduce an innovative input modality—3D VR sketches, to the 3D garment creation. We argue that using 3D VR sketches as input in AI generative models offers unique advantages over other methods. \textbf{Image input} struggles with freely creating 3D models from scratch, while \textbf{text input} is much less intuitive and precise in conveying spatial and geometric information compared to hand-drawn sketches. 
    
\subsection{2D Sketches or 3D Sketches}
    As sketches are natural form of computer-human interaction, some methods for 3D model generation from \textbf{2D sketches} have been proposed \cite{zhang2021sketch2model, guillard2021sketch2mesh, chen2023deep3dsketch, olsen2009sketch, bonnici2019sketch, lun20173d,li2018robust, wang20223d, zhong2020deep, gao2022sketchsampler, chen2023deep3dsketch+, zang2023deep3dsketch+2, zang2023deep3dsketch+}. However, researches have found that 2D sketches are inherently ambiguous and abstract, leading to issues with occlusion and information loss when the viewpoint is limited. Creating view-consistent 2D sketches by hand is also nearly impossible. In contrast, with the growing popularity in AR/VR devices as consumer electronics, \textbf{3D VR sketches} can convey more comprehensive information, such as accurately describing internal features of complex objects, and also easier for users to understand, which have been confirmed by previous research \cite{chen2024rapid}. In this research, we are the first to our knowledge to expand the 3D VR sketches in the field of 3D garment creation.

\section{KO3DClothes Dataset}
\label{sec:KO3DClothes}
    Given the limited availability of datasets for VR sketch-to-3D garment generation, we first create a novel dataset, KO3DClothes. To ensure realism and capture the nuances of human-created sketches, we opted for manual annotation by human volunteers rather than relying on synthetic data. Sketches, unlike simple edge detections, inherently reflect human intention and imprecision, and only through manual annotation can we accurately capture the errors and subtle variations characteristic of hand-drawn strokes.
    
\begin{figure}[htb]
\centering
\includegraphics[width=0.3\textwidth]{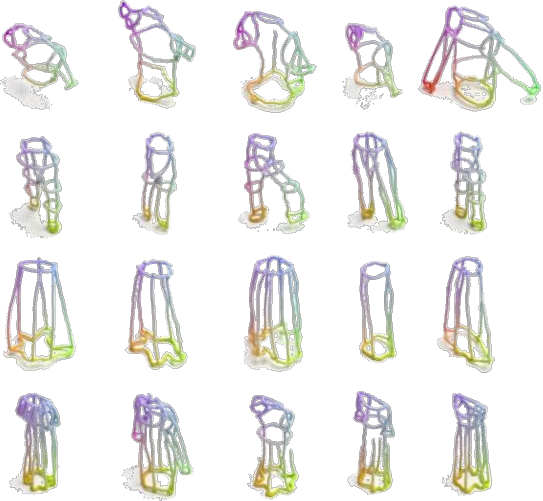}
\caption{The visualization of hand-drawn 3D sketch samples from KO3DClothes dataset.} \label{sketch}
\vspace{-0.5cm}
\end{figure}

    We invited 10 participants to create sketches on existing 3D models from the DeepFashion3D dataset \cite{zheng2021pamir}. During data collection, participants wore VR headsets and used controllers to draw sketches in a virtual environment, while employing dedicated software for the creation process. The system tracked the controllers' movements in real-time via VR. The data collection protocol was adapted from previous studies \cite{luo2021fine, chen2024rapid}. Specifically, participants were asked to draw 3D sketches around manually crafted 3D models within a predefined boundary box. Each participant’s strokes were recorded as a series of 3D coordinates, forming point cloud data that represents the 3D structure of the sketch. In each sample, the point cloud data was uniformly sampled with N=4096 points. Furthermore, the sketches were manually aligned to ensure consistent X, Y, and Z directions and positions, providing a standardized reference frame for accurate comparison and analysis. As a result, we obtained aligned and paired 3D sketches and 3D models. The 3D models used were selected from the high-quality models of the DeepFashion3D dataset \cite{zheng2021pamir}, consisting of 969 unique 3D shapes. Each 3D sketch-shape pair went through a quality control step by another participant. Fig. \ref{sketch} presents representative examples of the user-created 3D sketches of 3D clothes.

\section{Method}
\label{sec:Method}

\begin{figure*}[htb]
\centering
\includegraphics[width=0.8\textwidth]{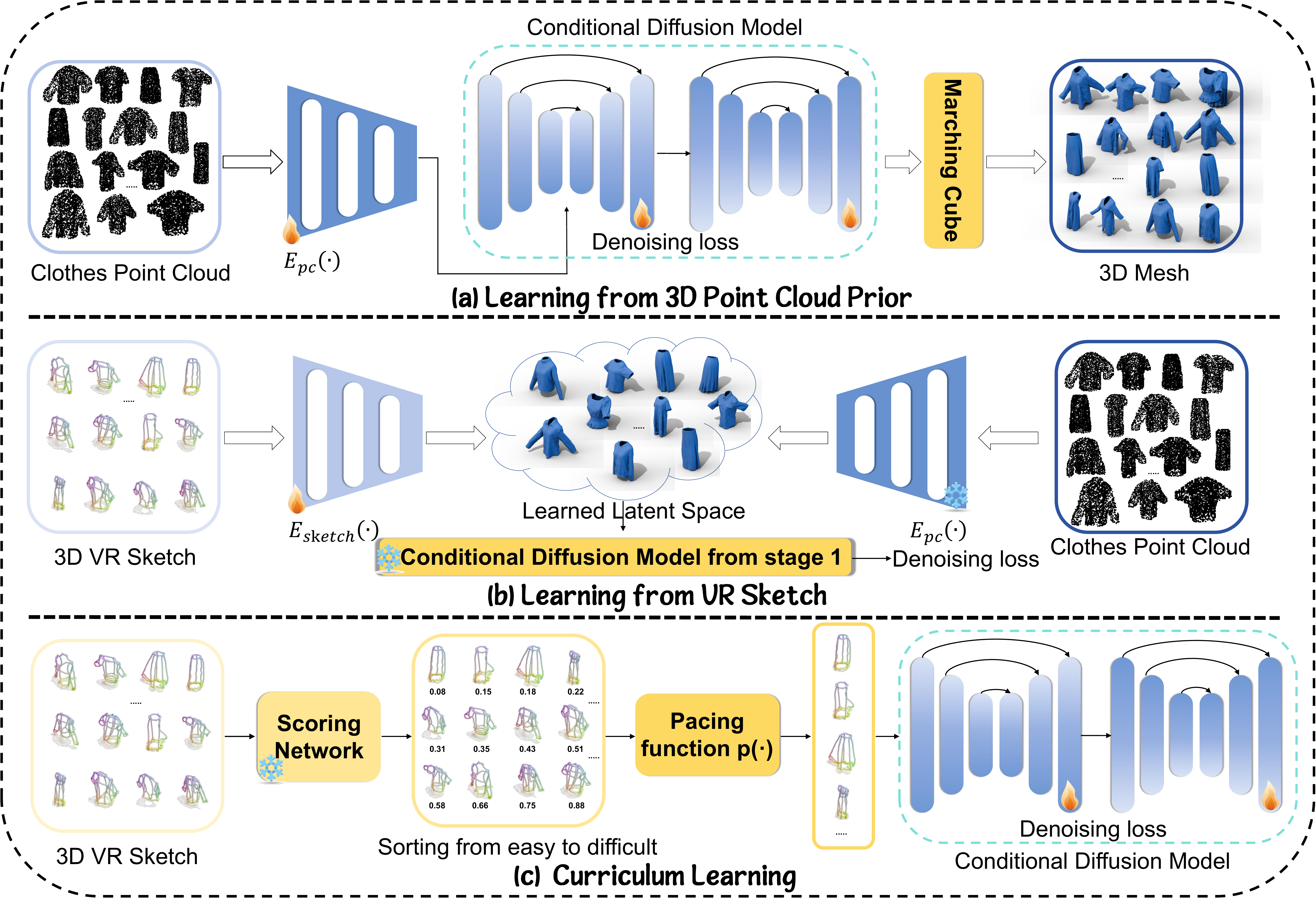}
\caption{The Overview of Deep3DVRSketch+. (a) Pre-training a conditional diffusion model by sampling ground truth (GT) point clouds. (b) Fine-tuning the sketch encoder to project sketches onto the diffusion manifold. (c) Curriculum learning leverages a limited set of sketch-shape pairs.} \label{Fig1}
\vspace{-0.5cm}
\end{figure*}


\subsection{Preliminary: Conditional 3D Diffusion Model}
\subsubsection{Principle}
    Our method employs a conditional diffusion model to generate 3D shapes, which have shown excellent performance in generating diverse and high-quality 3D models.  
    We train the diffusion model by reversing the noise diffusion process to sample from the target distribution. Given a sample \( z \), we gradually add Gaussian noise \( z_t \) to the sample based on a pre-defined variance schedule, with \( t \) ranging from 1 to \( T \), producing the corresponding \( z_t \). Next, we use a time-conditioned 3D UNet (denoted as \( \epsilon_\theta \)) to denoise the noise. Finally, the UNet generates a new 3D shape from the denoised Gaussian noise sample.


\subsubsection{Coarse-to-Fine Diffusion Network}
    To achieve high-fidelity 3D shape representations, we use high-resolution discrete signed distance fields (SDFs) to accurately model shape details. However, directly generating a dense SDF grid can incur substantial computational and memory costs due to its cubic complexity. To mitigate this computational burden while ensuring high-quality model output, we reference the method proposed in \cite{zheng2023locally} and design a two-stage diffusion framework, employing a self-conditioned continuous diffusion model. Specifically, the first stage uses a 5-layer UNet to generate a low-resolution 3D occupancy volume \( C \in R^{n \times n \times n} \), providing a coarse representation of the 3D shape. Next, the second stage constructs a high-resolution sparse volume \( F \in R^{N \times N \times N} \) using a 4-layer UNet, where an octree convolutional neural network is employed to handle the sparse voxel format of the SDF data. Both UNets are trained with a denoising loss \cite{ho2020denoising}, which is formulated as:
\begin{align}
    \begin{split}
     L(\theta)=\mathbb{E}_{z, \epsilon \sim N(0,1), t}\left[\left\|\epsilon-\epsilon_\theta\left(z_t, t, \{\mathbf{c}_i\}_{i=1}^{N}\right)\right\|^2\right]
    \end{split}
\end{align}
    where $\mathcal{N}(0,1)$ represents the Gaussian distribution, and $\{\mathbf{c}_i\}_{i=1}^{N}$ denotes the conditions required for the generative process. In the implementation, \( n \) is set to 32, and \( N \) is set to 128.

\subsubsection{Learning the Conditional Distribution}
    We use the conditional diffusion model as the core structure to generate 3D content, where the user-provided condition $\{\mathbf{c}_i\}_{i=1}^{N}$ is injected into the model to generate results based on the user's needs. Conditional features are extracted through a dedicated encoder, which converts condition data, such as point clouds, into 1024-dimensional feature embeddings $l$. These feature embeddings are then integrated into the UNet architecture using a multi-head cross-attention mechanism. Due to the limitations of the conditional input, the diffusion model may restrict the diversity of generated samples. We use classifier-free guidance \cite{ho2022classifier} to mitigate this issue.

\vspace{-0.3cm}
\subsection{The Proposed Multi-stage Training Steategy}
    In this section, we introduce our multi-stage training strategy. We first train a 3D diffusion generator that can generate high-quality shapes based on processed point cloud data, which is provided by the point cloud encoder $E_{pc}(\cdot)$. This process generates a feature encoding \( l \), which guides the diffusion model for fine-tuning and generates realistic 3D shapes. Then, in the 3D sketch mapping stage, we keep the diffusion generator unchanged and train the sketch encoder $E_{sketch}(\cdot)$ to map the input 3D sketch \( S_{vr} \) into the latent space \( Z \) from the first stage. It is important to note that while the output at this stage is close to the latent space, it still deviates from perfect alignment with the pre-trained input. Therefore, we introduce a joint fine-tuning stage, where both the encoder \( E(\cdot) \) and the diffusion generator are fine-tuned simultaneously to optimize the alignment and improve the quality of alignment between \( E_{pc}(\cdot) \) and \( E_{sketch}(\cdot) \). This multi-stage training approach helps us fully leverage the knowledge of the pre-trained model and has been proven to be crucial for improving the final output quality.

\subsubsection{Stage 1: Pre-Training the 3D Priors}
    In this stage, our primary goal is to train the conditional diffusion model to generate 3D shapes in the latent space, providing important prior information for the subsequent 3D sketch-to-3D model conversion. To achieve this, we use 3D point cloud data as conditional input and employ a supervised diffusion model for generation. Specifically, we use a pre-trained point cloud encoder, Uni3D \cite{zhou2023uni3d}, which plays a key role in encoding 3D point cloud data. This encoder \( E_{pc}(\cdot) \) converts the 3D point cloud into a latent encoding \( l^{+}_{pc} = E_{pc}(S_{pc}) \in \mathbb{R}^{1024} \) in the diffusion model. This approach allows the model to accurately capture the latent morphological features present in different 3D garment shape datasets and generate high-quality 3D shapes.

\subsubsection{Stage 2: 3D Sketch Mapping}
    In this stage, we input hand-drawn 3D sketches into the system and map them to the latent space \(Z\) defined in Stage 1, ensuring alignment between the sketches and the point cloud data. During this stage, we keep the diffusion model pre-trained in the first stage unchanged. Since 3D sketches are represented through point clouds, we design a Transformer-based point cloud encoder \( E_{sketch}(\cdot) \) to map the sketch \( S_{vr} \) into a latent encoding \( l^{+}_{vr} = E_{sketch}(S_{vr}) \in \mathbb{R}^{1024} \), with dimensions consistent with the Uni3D features from the previous stage. 
    

\subsubsection{Stage 3: Joint Fine-Tuning}
    We found that simply adjusting the 3D sketch encoder \( E_{sketch}(\cdot) \) does not ensure optimal alignment between the generated shapes and the sketch input. Inspired by certain approaches in image diffusion generation methods \cite{wang2022pretraining}, we decided to simultaneously fine-tune both the 3D sketch encoder \( E_{sketch}(\cdot) \) and the diffusion model to enhance spatial-semantic alignment. This strategy helps to fully leverage the knowledge of the pre-trained model and is crucial for improving the quality of the generated results. 
    
    Experimental results show that, although the fine-tuned model generates results of high-quality, there is still room for improvement, especially in model details, as shown in \ref{Ablation_results}. To address this challenge, we propose an adaptive curriculum learning strategy that focuses on improving the model's details.

\vspace{-0.3cm}
\subsection{Adaptive Curriculum Learning}
    Data scarcity and the complexity of 3D sketches are two major challenges we face. In our framework, we observed that when the training data consists of a limited number of hand-drawn 3D sketches and their corresponding 3D models, the network typically struggles to effectively generalize across a broader range of sketch styles and geometries when mapping these abstract sketches to the latent space and conditioning them. 
    Inspired by curriculum learning, we address these issues by simulating the way humans learn during the sketching process. Just as beginners in sketching typically start to learn with with simple and flexible shapes and gradually progress to more complex forms, we aim to apply this progressive learning strategy to our framework.

\subsubsection{Sample Difficulty Score}
    In curriculum learning, carefully selecting and ordering samples from simple to complex is crucial for effective progressive skill development. The selection of samples is based on their difficulty scores. Inspired by the curriculum DeepSDF \cite{duan2020curriculum}, we treat points with estimation errors as difficult samples, points with correct estimations as easy samples, and points with values between 0 and the true value as semi-difficult samples. We use the following difficulty scoring formula: 

\vspace{-0.5cm}
\begin{align}
    \begin{split}
     s = 1 + \alpha \, \text{sgn}(y) \, \text{sgn}(\bar{y} - y)
    \end{split}
\end{align}
    where \( y \) is the SDF value corresponding to the hand-drawn sketch, \( \bar{y} \) is the predicted SDF value, and \( \alpha \) controls the coefficients for difficult and semi-difficult samples. The function \( \text{sgn}(v) \) is defined as:  
    \[
    \text{sgn}(v) = 
    \begin{cases} 
    1 & \text{if } v \geq 0, \\
    -1 & \text{if } v < 0.
    \end{cases}
    \]

\subsubsection{Adaptive Curriculum}
    Unlike traditional manually designed curriculum learning, we adopt an adaptive curriculum learning strategy \cite{kong2021adaptive}. Specifically, we first use a pre-trained network to obtain the initial difficulty scores and, based on these scores, sort the initial dataset \( \Lambda \) in ascending order to form a sample pool \( \Lambda^{'} \). Then, we divide \( \Lambda^{'} \) into different mini-batches \( B=[B_1,..., B_m] \) and sequentially input them into the target network for training. Next, we design a pacing function \( p(\cdot) \), which is a monotonically increasing function that determines the rate at which we learn from simpler to more complex samples. Finally, at the end of the forward propagation, we update the difficulty scores and calculate the new sample pool \( \Lambda^{''} \). The difficulty score at the \( (k+1) \)-th position can be represented as:

\vspace{-0.4cm}
\begin{align}
    \begin{split}
        s_{k+1} = (1 - \beta)s_k + \beta s
    \end{split}
\end{align}
    where $k=\lfloor B_m/inv\rfloor $, \( \text{inv} \) controls the frequency of difficulty score updates, and \( \beta \) controls the speed at which the difficulty scores are updated.

\subsubsection{Pacing Function}
    To manage the pace at which the network learns from the samples, we need a monotonically increasing pacing function \( p(\cdot) \) to control the rate at which data is fed into the target network. This function can be expressed as: 

\vspace{-0.4cm}
\begin{align}
    \begin{split}
        p(i) = n \times \min(1, p_0 \times q^{\lfloor i/r_0 \rfloor })
    \end{split}
\end{align}

    where $n$ represents the number of samples, \( p_0 \) is the sample ratio in the initial step, \(q \) controls the speed of sample ratio growth, \( r_0 \) controls the frequency of sample ratio growth, and \( i \) is the current step. In actual training, we set \( p_0 \) to 0.2, \( q \) to 1.9, and \( r_0 \) to 1.

\section{Experiments}
\label{sec: Experiments and results}

\subsection{Implementation Details}
    In this study, we first pre-train the generative diffusion model on 3D shape dataset (DeepFashion3D) to enable high-quality synthesis (Stage 1). Subsequently, the model undergoes fine-tuning on a paired dataset of hand-drawn 3D sketches and 3D shapes with our KO3DClothes+ dataset (Stages 2-3). We split the KO3DClothes dataset into training and test sets in an 8:2 ratio. In the initial pre-training stage, we train the first UNet using the Adam optimizer \cite{kingma2014adam}, with a learning rate of 2e-4 for 800 epochs. For training the second UNet, we adopt the AdamW optimizer \cite{loshchilov2017decoupled}, adjusting the learning rate to 1e-4 and training for 500 epochs. During the 3D sketch mapping stage, we train the sketch encoder using the Adam optimizer with a learning rate of 2e-4 for 300 epochs. Finally, in the joint fine-tuning stage, we simultaneously train both the diffusion model and the 3D sketch encoder using the Adam optimizer for 300 epochs, maintaining a learning rate of 2e-4. The training process is conducted on Ascend 910b GPUs under Mindspeed framework. Following previous study \cite{chen2024rapid}, we use widely-used voxel IoU (Intersection over Union) and Chamfer Distance (CD) to evaluate the model performance.

\vspace{-0.3cm}
\subsection{Qualitative and Quantitative Assessment}
    To evaluate the performance of our method, we conducted comparison on several recent 3D reconstruction and generation methods based on VR sketches. 3DSketch2Shape \cite{luo20233d} is an early attempt of 3D sketch to 3D shape based on normalizing flow, our diffusion model exhibits better shape production capability. Deep3DVRSketch \cite{chen2024rapid} also use diffusion model, but this approach is initially designed for object 3D content creation, while our approach is specifically designed for garment creation. We use point-cloud 3D prior instead of image priors as in \cite{chen2024rapid} to better fit the complex nature of the clothes, which demonstrate better performance in the experiment. 
    

    
\begin{figure}[htb]
\centering
\includegraphics[width=0.48\textwidth]{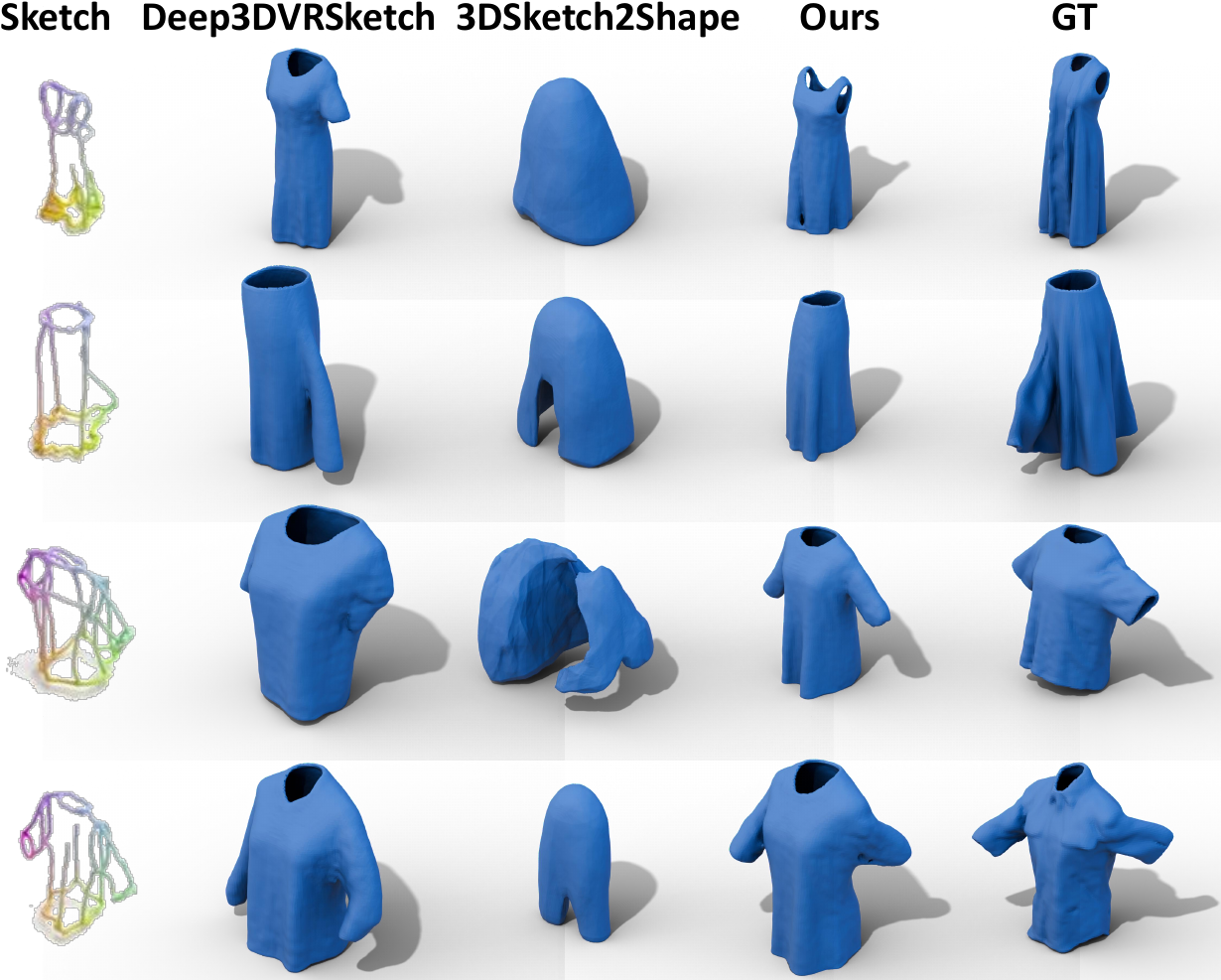}
\caption{Comparison with the existing state-of-the-art methods.} \label{exp1}
\vspace{-0.5cm}
\end{figure}

    We train each baseline model with the same KO3DClothes dataset. As shown in Table \ref{main_results}, our method performs well in shape accuracy. To view the shape quality, Fig. \ref{exp1} presents visualization of results. Our approach is capable of generating plausible and high-quality 3D garments. It's important to note that the primary focus of our experiment is capturing the \textbf{overall shape} of the garment (w/ or w/o sleeves, the length of the sleeves, etc.). Fine details like wrinkles and folds are not well-suited for, nor necessary to obtain through, 3D sketching. Leveraging the capabilities of numerous commercially available clothing simulators, we can readily generate dynamic garment with realistic clothing behavior with natural wrinkles across various poses from clothes simulations once the overall shape is established. 

\vspace{-0.5cm}
\begin{table}[h]
\centering
\caption{Quantitative Evaluation of the Real-World KO3DClothes Dataset}
\vspace{-0.2cm}
\label{main_results}
\resizebox{0.3\textwidth}{!}{
\begin{tabular}{ccc}                                                                                            \hline
    Methods &   IoU↑    &   CD↓ \\
     \hline
    3DSketch2Shape  & 0.3188    & 0.0820    \\
    Deep3DVRSketch   & \textbf{0.3252}   & 0.0606    \\
    \textbf{Ours}        & 0.3190     & \textbf{0.0597}  \\ \hline
\end{tabular}
}
\vspace{-0.6cm}
\end{table}

\subsection{User Study}
\subsubsection{3D Model Quality and Fidelity}
We evaluate the quality of generated results from user study. 
   Following previous research (Chen et al., 2023a), we used the widely adopted 5-point Mean Opinion Score (MOS) metric for evaluation. In this study, participants were asked to rate the generated 3D models based on two aspects, with scores from 1 to 5: Q1) The degree of fidelity between the generated 3D model and the input sketch; Q2) Participants' overall evaluation of the quality of the generated 3D model. We invited 15 designers to participate and presented them with 12 results generated by our algorithm for evaluation. Before the experiment began, we provided detailed explanations of the "fidelity" and "quality" rating criteria to ensure that participants had a consistent understanding of the evaluation standards. The average ratings of the experimental results are shown in Tab. \ref{User_Study1}. Compared to existing methods, our approach performed obtains higher user ratings, which further validating the advantages and effectiveness of our method in generating high-quality 3D garments.

\vspace{-0.5cm}
\begin{table}[h]
\centering
\caption{Mean Opinion Scores (1-5) from User Study}
\vspace{-0.2cm}
\label{User_Study1}
\resizebox{0.4\textwidth}{!}{
\begin{tabular}{ccc}                                                                                            \hline
    Methods &   (Q1): Fidelity    &   (Q2): Quality \\
     \hline
    3DSketch2Shape  & 1.2175 $\pm$ 0.2813    & 1.3480 $\pm$ 0.6109 \\
    Deep3DVRSketch   & 3.2824 $\pm$ 0.6563  & 3.3921 $\pm$ 0.5014  \\
    \textbf{Ours}        & \textbf{4.5925 $\pm$ 0.4821}     & \textbf{4.6225 $\pm$ 0.3428}  \\ \hline
\end{tabular}
}
\vspace{-0.2cm}
\end{table}

\subsubsection{Comparison with 2D Sketches}
    To verify the advantages of our method in terms of controllability, additional user experiments were conducted by recruiting a group of 15 3D designers from a 3D printing company. We asked the designers to draw their desired shape contours on a 2D plane and use baseline models Sketch2model \cite{zhang2021sketch2model}. The designers also described and sketched their desired 3D models and generated textured 3D shapes using our approach. Participants were asked to evaluate the controllability and usefulness of each method, which are crucial factors in the assessment of user interface usability and user experience \cite{oh2018lead, albert2022measuring, zang2024magic3dsketch}. Based on the settings in \cite{albert2022measuring, zang2024magic3dsketch}, we employed a 7-point Likert scale ranging from strongly disagree to strongly agree. The results are shown in Table \ref{User_Study2}. Compared to existing 2D-to-3D methods, participants gave our 3D-to-3D approach higher ratings in controllability and usefulness.

\vspace{-0.3cm}
\begin{table}[h]
\centering
\caption{Mean Opinion Scores (1-7) from User Study}
\vspace{-0.2cm}
\label{User_Study2}
\resizebox{0.4\textwidth}{!}{
\begin{tabular}{ccc}                                                                                            \hline
    Methods &   (Q1): Controllability    &   (Q2): Usefulness \\
     \hline
    Sketch2model  & 2.5233 $\pm$ 1.0717    & 2.4883 $\pm$ 1.3165 \\
    \textbf{Ours}        & \textbf{6.4533 $\pm$ 0.6258}     & \textbf{6.6217 $\pm$ 0.3684}  \\ \hline
\end{tabular}
}
\vspace{-0.6cm}
\end{table}

\subsection{Ablation Study}

\begin{figure}[htb]
\centering
\includegraphics[width=0.48\textwidth]{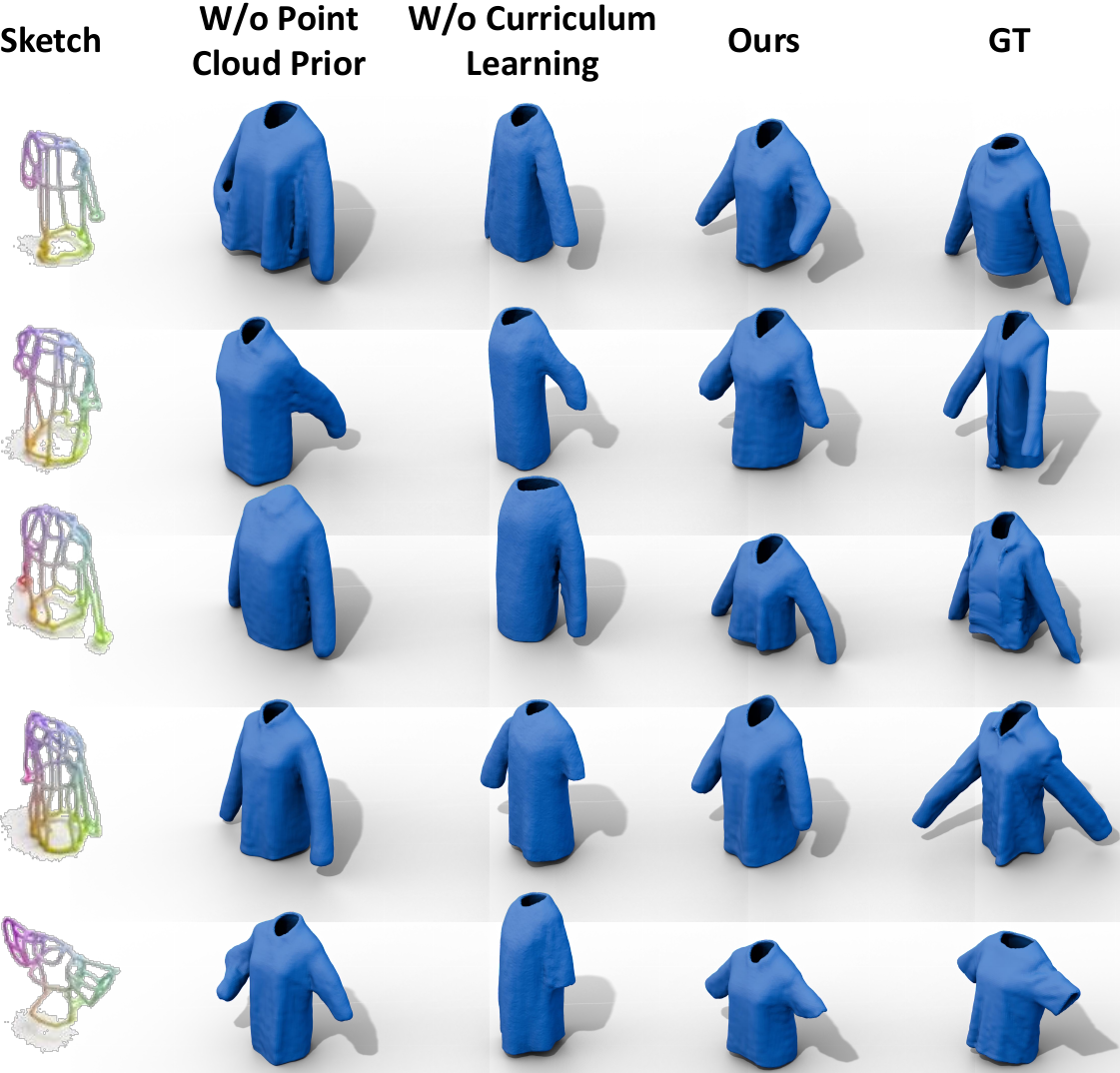}
\caption{Qualitative Evaluation for Ablation Studies.} \label{exp2}
\vspace{-0.3cm}
\end{figure}

    In the ablation study, we isolate the effectiveness of using shape priors and curriculum learning. We found that removing pretraining  with point cloud and directly training a conditional diffusion model on sketches (using the same network architecture) leads to a significant drop in performance. As shown in Fig. \ref{exp2} and Tab. \ref{Ablation_results}, models without pretraining fail to effectively learn to generate reasonable shapes. It is a effective way to overcome the challenge of limited annotated data in the sketch domain.

\vspace{-0.1cm}
\begin{table}[h]
\centering
\caption{Quantitative Evaluation of Ablation Study}
\vspace{-0.2cm}
\label{Ablation_results}
\resizebox{0.3\textwidth}{!}{
\begin{tabular}{cccccc}                                                                                            \hline
     &   IoU↑    &   CD↓   \\
     \hline
    w/o Point Cloud Prior  & 0.3178    & 0.0749    \\
    w/o Curriculum Learning   & 0.2680   & 0.0620   \\
    \textbf{Ours}        & \textbf{0.3190}     & \textbf{0.0597}   \\
     \hline
\end{tabular}
}
\vspace{-0.4cm}
\end{table}

    The curriculum learning method guides the model through a step-by-step process from simple to complex learning tasks, which significantly improves the model’s ability to adapt to diverse sketch styles as shown in Tab. \ref{Ablation_results}.

\vspace{-0.1cm}

\section{Conclusion}
As immersive consumer electronics continue to redefine how people interact with 3D content, our work takes a step toward democratizing digital fashion by making 3D garment design intuitive, expressive, and widely accessible. By harnessing the natural interaction affordances of AR/VR devices and the power of generative AI, we enable users to turn simple 3D sketches into detailed, wearable virtual garments without the need for professional tools or training. Our novel architecture and data collection efforts not only push the frontier of sketch-based modeling but also pave the way for broader adoption in consumer-facing applications, such as personalized avatars, virtual try-ons, and digital self-expression. Looking forward, we envision expanding this framework to real-time, interactive design experiences on everyday AR/VR devices — bringing the future of fashion creation to the fingertips of every user.

\bibliographystyle{IEEEtran}
\bibliography{main}

\vfill

\end{document}